\documentclass[10pt]{article}

\ifdefined\TMLRAnonym
  \usepackage{tmlr}
  \def\ReviewVersion{1} 
\else
  \usepackage[preprint]{tmlr}
\fi

\usepackage{graphicx}
\usepackage{amsmath}
\usepackage{amssymb}
\usepackage{booktabs}
\usepackage{xspace}
\usepackage[dvipsnames]{xcolor}
\usepackage[format=plain,labelformat=simple,labelsep=period,font=small]{caption}
\usepackage[font=footnotesize,skip=3pt,subrefformat=parens]{subcaption}
\usepackage[hyphens]{url}
\usepackage[shortlabels,inline]{enumitem}

\newlength{\spalte}
\usepackage[T1]{fontenc}
\usepackage{cuted}     
\usepackage{microtype}
\usepackage{multirow}
\usepackage{bm}

\newcommand{\qwen}{Qwen\xspace}
\newcommand{\kontext}{Kontext\xspace}
\newcommand{\longcat}{LongCat\xspace}
\newcommand{\pcover}{\ensuremath{P_\text{cover}}\xspace}
\newcommand{\preplace}{\ensuremath{P_\text{replace}}\xspace}
\newcommand{\ci}[2]{\,{\scriptsize[#1,\,#2]}}
\newcommand{\degr}{\ensuremath{^\circ}}

\providecommand{\spalte}{\linewidth}



\makeatletter
\DeclareRobustCommand\onedot{\futurelet\@let@token\@onedot}
\def\@onedot{\ifx\@let@token.\else.\null\fi\xspace}
 
\def\ie{\emph{i.e}\onedot} 
 
 \def\vs{\emph{vs}\onedot}
 
\makeatother

\renewcommand{\cite}{\citep}
\renewenvironment{strip}{\begin{figure}[!ht]}{\end{figure}}
\newcommand{\supplementanfang}{\clearpage}

\definecolor{cvprblue}{rgb}{0.21,0.49,0.74}
\usepackage[breaklinks,colorlinks,allcolors=cvprblue]{hyperref}
\ifdefined\TMLRAnonym\else
  \hypersetup{pdftitle={The Camera Inside the Editor: Reading the Implicit Camera of Image Editors with Painted Calibration Patterns},
              pdfauthor={Sebastian R\"uckerl}}
\fi
\usepackage[capitalize]{cleveref}
\crefname{section}{Sec.}{Secs.}
\Crefname{section}{Section}{Sections}
\Crefname{table}{Table}{Tables}
\crefname{table}{Tab.}{Tabs.}

\title{The Camera Inside the Editor:\\Reading the Implicit Camera of Image Editors\\with Painted Calibration Patterns}

\author{\name Sebastian R\"uckerl \email sebastian@rückerl.com\\
      \addr University of Hagen, Germany}

\def\month{MM}
\def\year{YYYY}
\def\openreview{\url{https://openreview.net/forum?id=XXXX}}

\begin{document}
\maketitle
\begin{abstract}
Instruction-based image editors insert objects, restyle scenes and render new viewpoints, but it is unknown which camera they assume when they paint into a photograph.
Asked to cover the floor with a checkerboard, an editor paints projective structure from which classical vanishing-point geometry reads pitch, roll, focal length, yaw and, on renders, the principal point, without any training.
Unlike a calibrator such as GeoCalib, which estimates the camera of an image, this isolates the camera under which the editor paints.
On 120 rendered cameras with exact ground truth, Qwen-Image-Edit-2511 paints tile edges that meet their vanishing points within 0.26\degr, and its implicit camera matches the true one to 0.8\degr{} in pitch and 6\% in focal length, more accurately than GeoCalib except in roll.
Asked to draw the horizon or mark a vanishing point instead, the editor fails, so this knowledge is revealed by painting and not by the explicit tasks we tried.
The implicit camera has two priors: roll is pulled towards level (slope 0.71), and telephoto perspective towards a default of about 30\,mm, which roughly matches the camera the models paint without any scene.
For \qwen, the priors do not grow when blur removes four fifths of the line evidence.
They are stronger on real photographs, and on NYUv2 a shorter wording of the task removes the difference for roll.
On photographs from a 24--240\,mm zoom lens the painted perspective grows with only 0.62 of the lens's slope, while GeoCalib and MoGe-2 saturate at about 52 and 42\,mm.
FLUX.1 Kontext and LongCat-Image-Edit are pulled much harder.
Finally, from a level camera a camera-control LoRA executes pose commands at only 50--70\% of their strength, and a board painted into its output agrees with the camera it produced.
\end{abstract}

\begin{strip}
\centering
\includegraphics[width=\textwidth]{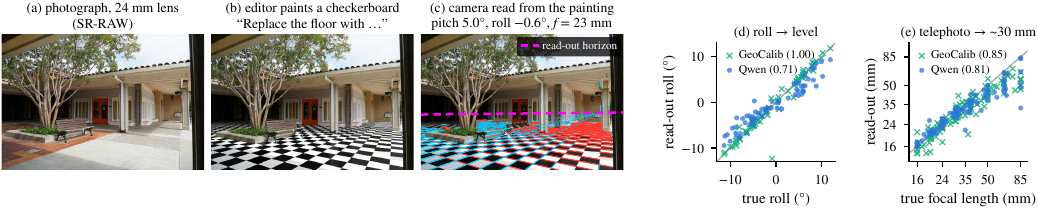}
\captionof{figure}{\textbf{Reading the camera an editor assumes.}
(a)~A real photograph from SR-RAW~\cite{zhang2019zoom}, taken with a 24\,mm lens.
(b)~Asked to replace the floor with a checkerboard, Qwen-Image-Edit-2511 paints one in perspective.
(c)~The painted tile edges (red, blue) meet in vanishing points, from which we read the camera the editor assumed, with its horizon (magenta), pitch, roll and focal length (here 23\,mm).
(d,\,e)~On 120 rendered cameras, this camera is pulled towards a default, with roll pulled towards level and long focal lengths towards about 30\,mm.}
\label{fig:teaser}
\vspace{0.3em}
\end{strip}

\section{Introduction}
\label{sec:intro}

Instruction-based image editors have become general-purpose tools for manipulating photographs: they insert people and objects, restyle scenes, synthesise training data, act as zero-shot perception models~\cite{liu2026opensource,gabeur2026visionbanana,di2026probegen} and render new viewpoints on command~\cite{fal2026multipleangles,shen2026cameraeditor}.
Every such edit implies a camera, since an inserted chair has a size and a perspective and a painted floor pattern converges somewhere.
However, it is unknown \emph{which} camera an editor assumes and how it errs.
Existing evaluations judge edits from the outside, by the plausibility of object sizes~\cite{li2026genscale} or by the viewpoint reconstructed from the output~\cite{xiao2026spatialedit}, and none of them measures the camera the editor itself uses.

We make this implicit camera visible with a calibration pattern (\cref{fig:teaser}).
We ask the editor to cover the floor with a checkerboard, whose image is fully determined by the camera, and read the painting with textbook geometry~\cite{caprile1990vanishing,hartley2004multiple}.
The two families of tile edges give horizon, roll, pitch and focal length, painted vertical poles add the principal point, and since the editor aligns the board with the walls, the board also gives the yaw relative to the room.
The editor remains a black box, and no weights, features or training are needed.
This extends painted probes, established for illumination and colour~\cite{phongthawee2024diffusionlight,chang2025gcc,giroux2026sheddinglight}, to camera geometry.
Why not simply run a camera calibrator such as GeoCalib~\cite{veicht2024geocalib} on the edited image? A calibrator estimates the camera of an image as a whole, and most of an edited image is the unchanged input. The camera under which the editor paints its addition can differ from it, and only the painted structure reveals it. Our aim is therefore not a better calibrator but a measurement of the editor.
To know what ``correct'' means, we render a catalog of cameras with exact ground truth, together with controls that separate the editor from the read-out and from copying, and we change the camera of real photographs in a known way.

The implicit camera of Qwen-Image-Edit-2511 turns out to be precise and systematically biased.
Its painted lines meet their vanishing points within 0.26\degr, and on renders its camera is more accurate than GeoCalib in horizon, pitch and focal length, though not in roll. On real photographs GeoCalib is more accurate in roll and focal length, which is where the editor's priors act.
This knowledge is implicit, because when asked to draw the horizon, the editor draws a line through the middle of the image.
Roll is pulled towards level and telephoto perspective towards a default of about 30\,mm, which roughly matches the camera the models paint when there is no scene to follow.
For \qwen, these priors do not grow when most of the line evidence is blurred away.
They are stronger on real photographs, and on NYUv2 a shorter wording of the task removes the difference for roll.
With a real zoom lens the telephoto prior persists, and learned calibrators saturate even earlier.
Two further editors, FLUX.1 Kontext~\cite{bfl2025kontext} and LongCat-Image-Edit~\cite{longcat2025image}, are pulled much harder.
Finally, a camera-control LoRA executes pose commands at reduced strength, and a board painted into its output matches the camera it actually produced.

\noindent\textbf{Contributions.}
(1)~Painted calibration probes, a training-free read-out of the full implicit camera of image editors, validated against a rendered oracle.
(2)~A camera catalog with exact ground truth and controls for read-out, copying, image quality and domain, complemented by real photographs with exactly varied cameras and by optical zoom sequences.
(3)~A characterisation of the implicit camera of Qwen-Image-Edit-2511 in depth and of two further editors on the catalog (accuracy, projective fidelity, implicit \vs explicit knowledge, and two priors that point to a default camera and do not grow with line blur), and of the camera that pose commands write.

\section{Related Work}
\label{sec:related}

\noindent\textbf{Single-image calibration.}
Two orthogonal families of parallel lines determine the horizon and the focal length, a third family the principal point~\cite{caprile1990vanishing,hartley2004multiple}.
In real scenes, the difficulty lies in finding reliable lines.
Learned calibrators regress gravity and field of view.
Perspective Fields predict per-pixel up-vectors and latitudes~\cite{jin2023perspective}, GeoCalib combines learned priors with geometric optimisation~\cite{veicht2024geocalib}, AnyCalib regresses pixel rays~\cite{tiradogarin2025anycalib}, and diffusion-based calibrators fine-tune Stable Diffusion to output dense incidence maps or camera images from which intrinsics are fitted~\cite{he2025diffcalib,deng2025dmcalib}.
Other work estimates horizon and camera height from objects of known size~\cite{hoiem2008putting,zhu2020metrology,lee2023scalefield} or inpaints people as metric landmarks, given the intrinsics~\cite{zhao2024mfh}.
Our approach needs neither training nor known intrinsics, because the editor supplies the calibration target and classical geometry reads it.

\noindent\textbf{What generative models know.}
Small LoRAs extract depth, normals, albedo and shading from generators~\cite{du2023generative}, linear probes find scene geometry in diffusion features~\cite{chen2023beyond}, foundation models have been probed for 3D awareness~\cite{elbanani2024probing}, and instruction editors serve as zero-shot estimators of depth, normals and segmentation~\cite{liu2026opensource,gabeur2026visionbanana,di2026probegen}.
None of these asks for camera parameters.
Conversely, generated images violate projective geometry~\cite{sarkar2024shadows,okumura2026controlvp}, which has motivated perspective losses during training~\cite{upadhyay2023enhancing}.
We show that an editor asked to paint a known structure is projectively precise, and we measure \emph{which} camera it encodes.

\noindent\textbf{Painted probes.}
Painting a known object and reading it out is established for photometry: a chrome ball for illumination~\cite{phongthawee2024diffusionlight}, a colour checker for white balance~\cite{chang2025gcc}, and grey spheres to benchmark the lighting understanding of editors~\cite{giroux2026sheddinglight}. We extend the principle to camera geometry.

\noindent\textbf{Camera control.}
Text-to-image models rarely execute camera settings given in the prompt~\cite{yuan2025generative}, and their backbones qualitatively prefer eye-level views~\cite{lu2026viewpoint}.
Dedicated adapters and models add explicit control of the viewpoint~\cite{fal2026multipleangles,shen2026cameraeditor,liao2026puffin}.
SpatialEdit benchmarks such commands by reconstructing the produced viewpoint~\cite{xiao2026spatialedit}, and GenScale judges relative object sizes with a vision--language model~\cite{li2026genscale}.
These works measure the camera an edit \emph{produces}, whereas we read the camera an editor \emph{assumes} when it paints, also in generated views.

\section{Reading the Implicit Camera}
\label{sec:method}

\noindent\textbf{Painted probes.}
We ask the editor to add a structure whose image is determined by the camera up to its own placement.
Our standard probe \pcover{} asks the editor to cover the entire visible floor with a large black-and-white checkerboard of square tiles that lies flat on the floor, keeping everything else unchanged.
\Cref{sec:prompt} shows that the shorter \preplace{} = \emph{``Replace the floor with a black-and-white checkerboard floor.''} is the better probe, and we report results for both.
A second probe asks for six thin vertical magenta poles standing on the floor (all prompts in the supplement).
The editor is treated as a black box, and we never access its weights or features.

\noindent\textbf{Locating the painted structure.}
Because editors slightly rescale and shift their output, we register the edited image to the input with ORB features~\cite{rublee2011orb} and a RANSAC~\cite{fischler1981ransac} similarity transform and map all detections into input pixel coordinates, where the ground truth is defined.
Weakly supported registrations far from the identity are discarded (supplement).
The painted region is the largest connected area in which the edit differs strongly from the input and has high local contrast.
Inside it, LSD~\cite{grompone2010lsd}, run at full resolution, detects line segments longer than 2\% of the image height.

\noindent\textbf{Vanishing points and the floor pair.}
Up to three segment families are extracted one after another by RANSAC on vanishing points, with 1500 hypotheses from segment pairs.
A segment is an inlier if it points to the hypothesis within 1.5\degr, and the inlier set is refined by the smallest singular vector of the length-weighted matrix of homogeneous line coordinates.
The two floor families are chosen among pairs of families with at least 8 segments each.
At least 90\% of the pair's segments must lie below its own vanishing line, since an upright camera sees the floor below the horizon, and the mean image directions of the two families must differ by at least 10\degr.
Among the admissible pairs, we keep the one with the most segments.
These thresholds do not affect our conclusions (supplement).

\noindent\textbf{Camera read-out.}
Let $\mathbf{v}_1, \mathbf{v}_2$ be the homogeneous floor vanishing points.
The horizon is the vanishing line $\mathbf{l} = \mathbf{v}_1 \times \mathbf{v}_2$, and its angle to the image rows is the roll.
With the principal point $\mathbf{p}$ at the image centre and square pixels, orthogonality of the two painted families gives
\begin{equation}
  f^2 = -(\bar{\mathbf{v}}_1 - \mathbf{p})^\top (\bar{\mathbf{v}}_2 - \mathbf{p}),
  \qquad
  \theta = \arctan\frac{y_h - p_y}{f},
  \label{eq:f}
\end{equation}
where $\bar{\mathbf{v}}_i$ are the inhomogeneous points, $y_h$ is the horizon height at the image centre and $\theta$ the pitch.
We use $f$ only if both vanishing points lie within 10 image heights of the centre.
If the editor paints the board parallel to the image border, one vanishing point recedes to infinity and $f$ is undefined.
Because this bound also limits the focal lengths that can be read, we instead bound the vanishing points of the zoom photographs (\cref{sec:prompt}) at 10 read-out focal lengths, which an exact board passes up to 240\,mm.
On all other data, the choice hardly matters (supplement).
Painted poles supply the vertical vanishing point $\mathbf{v}_z$ (robust intersection of the pole axes).
The principal point is then the orthocentre of the triangle $(\bar{\mathbf{v}}_1, \bar{\mathbf{v}}_2, \bar{\mathbf{v}}_z)$~\cite{caprile1990vanishing}, and the focal length also follows from $f^2 = d(\mathbf{p}, \mathbf{l})\,\lVert \bar{\mathbf{v}}_z - \mathbf{p} \rVert$.
Because the editor aligns the board with the walls (\cref{sec:knows}), back-projecting $\mathbf{v}_1$ with the read-out camera, $\mathbf{d} \propto R^\top K^{-1}\mathbf{v}_1$, gives the camera's yaw relative to the room (modulo 90\degr).
Under the \emph{true} camera we back-project both painted families to measure the painting itself in 3D: its tilt against the floor, the angle between the families and its alignment with the room axes.

\noindent\textbf{Validating the read-out.}
To separate errors of the editor from errors of our measurement, we render a physical 0.5\,m checkerboard into every catalog scene (``oracle'') and run the identical pipeline.
On the 120 cameras of \cref{sec:benchmark} it recovers the correct tile size and the camera almost exactly, with the horizon to 0.003 image heights and the focal length to 0.7\% (\cref{tab:accuracy}).
Deviations measured on edited images are therefore not errors of the geometric read-out, although which paintings are readable can select the sample (\cref{sec:discussion}).

\section{Benchmark}
\label{sec:benchmark}

\noindent\textbf{Camera catalog.}
We render three Blender demo scenes (a cluttered classroom~\cite{seux_classroom}, the Barcelona Pavilion~\cite{cheggour_pavillion} and a modern apartment~\cite{dellatommasa_italianflat}) from 44 cameras each with Cycles~\cite{blender2026} at $1024\times768$ pixels.
Set~F is a focal sweep from 16 to 85\,mm (eight lenses) at two positions with pitch $-15$\degr{} and no roll.
Set~Z contains 24 random cameras with focal lengths of 16--85\,mm, pitch from $-35$\degr{} to $+5$\degr{} and roll within $\pm$12\degr.
Set~S has four level cameras with vertical lens shift (details in the supplement).
Unless noted, we evaluate on F$\,\cup\,$Z (120 cameras).
Set~S, whose principal point is off-centre, is evaluated separately.
Focal lengths are 35\,mm equivalents.

\noindent\textbf{Controls.}
Every camera is rendered (i)~with the original floor, (ii)~with a uniform grey floor material, which removes planks, rugs and texture (the pavilion's slab joints remain, as they are geometry), and (iii)~with a physically rendered checkerboard (the oracle).
Further controls replicate the manipulations we apply to photographs, with an exactly updated camera: a \emph{blur ladder} (down- and up-sampling to effective widths down to 64\,px), \emph{digital zoom} (centre crops of the 24\,mm render) and \emph{rotation} with cropping, also with a \emph{photo look} (noise, colour cast, JPEG).
The editor also restyles set~F as cartoons.

\noindent\textbf{Real photographs.}
From the NYUv2 test set~\cite{silberman2012nyu} we fit the floor plane to metric depth and floor labels by RANSAC, which yields the horizon and camera height, and keep 40 images with enough floor.
We change their camera synthetically but exactly, by rotating them by $\pm$6\degr{} and $\pm$12\degr{} with a centre crop that removes black corners, and by centre crops that correspond to 51--102\,mm instead of the original 34\,mm (160 images each, details in the supplement).
For optical zoom we use SR-RAW~\cite{zhang2019zoom}, sequences taken with a 24--240\,mm zoom lens from one standpoint, of which 36 show a floor or ground plane at three or more focal lengths (35 outdoors).
We measure the focal-length ratio between frames by registration and anchor it at the EXIF focal length of the widest frame (196 frames).
Crops of the widest frame with the field of view of each longer frame serve as digital twins, and rotating the widest frame by $\pm$6\degr{} and $\pm$12\degr{} gives a second roll test (180 frames with the unrotated ones, details in the supplement).

\begin{figure*}[t]
\centering
\includegraphics[width=\textwidth]{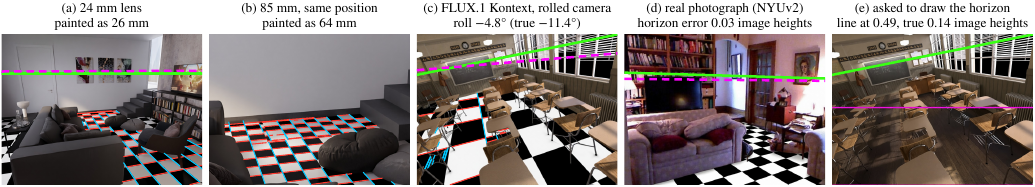}
\caption{\textbf{Qualitative results} (\qwen{} unless noted). Green: true horizon, dashed magenta: horizon read from the painted board, red/blue: detected tile edges.
(a,\,b)~The same position at 24 and 85\,mm. The telephoto board is painted too wide (64\,mm).
(c)~On a rolled camera, the board painted by \kontext{} is read as roll $-4.8$\degr{} (\qwen: $-10.9$\degr, truth: $-11.4$\degr).
(d)~On a real photograph the painted board recovers the horizon to 0.03 image heights, which is the median over NYUv2.
(e)~Asked to draw the horizon, the editor draws a level line (thin magenta) through the middle of the image.}
\label{fig:qualitative}
\end{figure*}

\noindent\textbf{Editors and baselines.}
Our main editor is Qwen-Image-Edit-2511~\cite{qwen2025edit2511,wu2025qwenimage} (``\qwen'') with its 4-step Lightning distillation~\cite{lightx2v2025edit2511lightning}, which gives a similar focal slope as the full model in the one scene we compared (0.67 \vs 0.66, supplement), and three variants (seeds) per image.
We compare FLUX.1 Kontext [dev]~\cite{bfl2025kontext} (``\kontext'', one or two variants per image) and, on the catalog, LongCat-Image-Edit~\cite{longcat2025image} (``\longcat'', one variant).
The default-camera experiment also includes OmniGen2~\cite{wu2025omnigen2} and the text-to-image model Krea~2 Turbo~\cite{krea2026turbo}.
As baselines we run GeoCalib~\cite{veicht2024geocalib} (gravity and focal length) and MoGe-2~\cite{wang2025moge2} (focal length) on the same inputs as the editor, and classical vanishing points detected on the input with our line pipeline.

\noindent\textbf{Metrics.}
Per image we take the median over variants.
We report the horizon error in image heights at the image centre, pitch and roll errors in degrees, and the focal error $|\log(\hat f/f)|$.
A prior is measured as the Theil--Sen slope~\cite{sen1968estimates} of read-out against truth, for pitch and roll on set~Z (where both vary) and for the focal length log--log on F$\,\cup\,$Z.
A slope of 1 means that the implicit camera follows the true one, and 0 means that it is fixed.
For the focal fit $\log\hat f = a\log f + b$ we also report the fixed point $f_0 = \exp(b/(1-a))$, towards which estimates are pulled.
Brackets give 95\% bootstrap confidence intervals over images~\cite{efron1979bootstrap} (over sequences for rotated SR-RAW frames, which share their unrotated reference).
Comparisons between conditions are paired on identical images.

\section{What the Editor Knows}
\label{sec:knows}

\begin{figure}[t]
\centering
\includegraphics[width=\spalte]{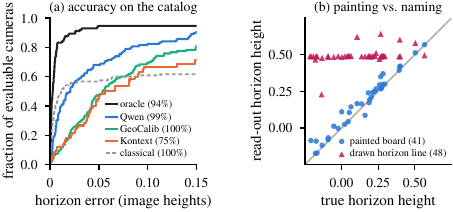}
\caption{(a)~\textbf{Horizon error} on the catalog (neutral floor, F\,$\cup$\,Z), cumulative over the evaluable cameras (share in parentheses). \kontext{} is evaluated in the classroom and the pavilion, the classical baseline on the original floor.
(b)~\textbf{Painting \vs naming} (set~Z, classroom and pavilion, original floor).
The horizon read from the painted board follows the truth, while the line the editor draws when asked stays at mid-height.}
\label{fig:accuracy}
\end{figure}

\begin{table}[t]
\centering
\caption{\textbf{Accuracy of the implicit camera} on the 120 catalog cameras (F\,$\cup$\,Z), medians over images: horizon error (image heights), pitch and roll errors (degrees) and focal error. The $f$~slope is the Theil--Sen slope of $\log\hat f$ on $\log f$, and $\perp$ is the deviation of the angle between the painted families in 3D from 90\degr.
Inputs have the neutral floor unless noted, and baselines see the same inputs as the probe.
$^\dagger$Classroom and pavilion (80 cameras), two variants on set~F and one on set~Z.
$^\S$One variant, board with a valid registration on 58 of the 120 cameras.
$^\ddagger$On the original-floor input, where $f$ exists for 57\% of the images. Horizon error 0.420 in the classroom and 0.008 and 0.005 in the other scenes.}
\label{tab:accuracy}
\footnotesize
\setlength{\tabcolsep}{3.1pt}
\begin{tabular}{@{}lcccccc@{}}
\toprule
 & Horizon & Pitch & Roll & $|\log f|$ & $f$ slope & $\perp$ \\
\midrule
Rendered board (oracle) & 0.003 & 0.1 & 0.1 & 0.007 & 0.99 & 0.3 \\
\midrule
\qwen{} probe & 0.021 & 0.8 & 0.7 & 0.060 & 0.81 & 2.0 \\
\quad original floor & 0.022 & 0.8 & 0.7 & 0.054 & 0.82 & 2.0 \\
\kontext{} probe$^\dagger$ & 0.062 & 2.1 & 2.5 & 0.226 & 0.43 & 6.8 \\
\longcat{} probe$^\S$ & 0.104 & 3.6 & 3.1 & 0.143 & 0.34 & 5.8 \\
\midrule
GeoCalib~\cite{veicht2024geocalib} & 0.054 & 1.6 & 0.2 & 0.094 & 0.85 & -- \\
\quad original floor & 0.046 & 1.5 & 0.2 & 0.073 & 0.92 & -- \\
MoGe-2~\cite{wang2025moge2} & -- & -- & -- & 0.123 & 0.60 & -- \\
Classical VPs$^\ddagger$ & 0.011 & -- & -- & 0.010 & 1.00 & -- \\
 
\bottomrule
\end{tabular}
\end{table}

\begin{figure*}[t]
\centering
\includegraphics[width=\textwidth]{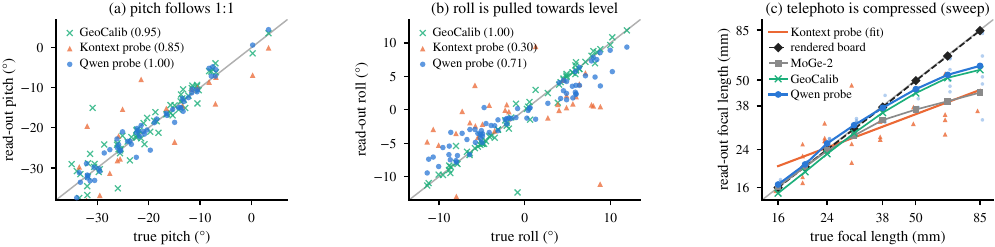}
\caption{\textbf{The implicit camera on the catalog} (neutral floor, Theil--Sen slopes in parentheses).
(a)~Pitch is read 1:1.
(b)~Roll is compressed towards level by both editors shown, but not by GeoCalib.
(c)~Focal sweep (set F, pitch $-15$\degr): the painted perspective is exact up to about 40\,mm and too wide beyond. Lines connect medians over scenes and positions (small dots: \qwen{} images). \kontext{}, with one to four valid images per lens, is shown with all images and a log--log fit.}
\label{fig:priors}
\end{figure*}

\begin{table*}[t]
\centering
\caption{\textbf{Priors of the implicit camera}: Theil--Sen slopes of read-out against truth (1: follows the camera, 0: fixed) with 95\% bootstrap intervals.
Renders have the neutral floor. Pitch and roll are measured on set~Z, focal length (log--log) on F\,$\cup$\,Z, and $f_0$ is the fixed point of the focal fit.
NYUv2: 40 images (\kontext: 20), rotated by $\pm$6\degr{} and $\pm$12\degr{} (roll) or centre-cropped to 34--102\,mm (focal length).
SR-RAW: roll on the widest frames of 36 sequences rotated by $\pm$6\degr{} and $\pm$12\degr{} (change against the unrotated frame, intervals over sequences), focal length on 196 optically zoomed frames (24--245\,mm) with vanishing points within 10 read-out focal lengths (\cref{sec:method}). The oracle row projects an exact board into the same frames.
$^\dagger$Classroom and pavilion only.
$^\S$One variant per image, 58 of 120 renders evaluable.
$^\ast$First variant per image, focal length on renders from set~Z, roll on NYUv2 from 20 base images (80 rotations), paired on identical images.}
\label{tab:priors}
\footnotesize
\setlength{\tabcolsep}{2.6pt}
\resizebox{\textwidth}{!}{%
\begin{tabular}{@{}lcccccccccc@{}}
\toprule
 & \multicolumn{4}{c}{Renders (catalog)} & & \multicolumn{2}{c}{NYUv2 photographs} & & \multicolumn{2}{c}{SR-RAW photographs} \\
\cmidrule(lr){2-5}\cmidrule(lr){7-8}\cmidrule(l){10-11}
 & Pitch & Roll & Focal length & $f_0$ & & Roll & Focal length & & Roll & Focal length \\
\midrule
Rendered board (oracle) & 1.01 & 1.00 & 0.99 & -- & & -- & -- & & -- & 1.00 \\
\qwen{} probe, \pcover & 1.00\,{\scriptsize[0.95,1.05]} & 0.71\,{\scriptsize[0.63,0.78]} & 0.81\,{\scriptsize[0.75,0.87]} & 33\,mm & & 0.65\,{\scriptsize[0.60,0.69]} & 0.23\,{\scriptsize[0.09,0.34]} & & 0.69\,{\scriptsize[0.59,0.77]} & 0.52\,{\scriptsize[0.37,0.73]} \\
\kontext{} probe, \pcover$^\dagger$ & 0.85\,{\scriptsize[0.58,1.17]} & 0.30\,{\scriptsize[0.09,0.45]} & 0.43\,{\scriptsize[0.27,0.66]} & 25\,mm & & 0.18\,{\scriptsize[0.07,0.33]} & $-$0.02\,{\scriptsize[$-$0.39,0.32]} & & -- & 0.36\,{\scriptsize[0.18,0.57]} \\
\longcat{} probe, \pcover$^\S$ & 0.84\,{\scriptsize[0.66,0.94]} & 0.25\,{\scriptsize[0.13,0.36]} & 0.34\,{\scriptsize[0.22,0.51]} & 28\,mm & & -- & -- & & -- & -- \\
GeoCalib~\cite{veicht2024geocalib} & 0.95\,{\scriptsize[0.87,1.04]} & 1.00\,{\scriptsize[0.98,1.02]} & 0.85\,{\scriptsize[0.79,0.90]} & 20\,mm & & 1.00\,{\scriptsize[0.98,1.01]} & 0.59\,{\scriptsize[0.53,0.68]} & & 1.00\,{\scriptsize[0.99,1.01]} & 0.43\,{\scriptsize[0.37,0.49]} \\
MoGe-2~\cite{wang2025moge2} & -- & -- & 0.60\,{\scriptsize[0.56,0.65]} & 26\,mm & & -- & 0.85\,{\scriptsize[0.81,0.90]} & & -- & 0.36\,{\scriptsize[0.33,0.40]} \\
\midrule
\qwen, \pcover$^\ast$ & -- & 0.76\,{\scriptsize[0.68,0.84]} & 0.76\,{\scriptsize[0.64,0.86]} & -- & & 0.63\,{\scriptsize[0.59,0.69]} & 0.27\,{\scriptsize[0.14,0.37]} & & 0.65\,{\scriptsize[0.55,0.74]} & 0.56\,{\scriptsize[0.40,0.73]} \\
\qwen, \preplace$^\ast$ & -- & 0.78\,{\scriptsize[0.71,0.85]} & 0.88\,{\scriptsize[0.80,0.97]} & -- & & 0.78\,{\scriptsize[0.71,0.83]} & 0.55\,{\scriptsize[0.44,0.66]} & & 0.83\,{\scriptsize[0.75,0.90]} & 0.62\,{\scriptsize[0.52,0.77]} \\
\quad paired difference & -- & $+$0.03\,{\scriptsize[$-$0.05,$+$0.10]} & $+$0.14\,{\scriptsize[$+$0.06,$+$0.25]} & -- & & $+$0.15\,{\scriptsize[$+$0.07,$+$0.20]} & $+$0.29\,{\scriptsize[$+$0.15,$+$0.43]} & & $+$0.18\,{\scriptsize[$+$0.07,$+$0.30]} & $+$0.02\,{\scriptsize[$-$0.12,$+$0.15]} \\
 
\bottomrule
\end{tabular}}
\end{table*}

\noindent\textbf{Accuracy.}
Read from the painted board, \qwen's implicit camera matches the true one to 0.021 image heights in horizon, 0.8\degr{} in pitch, 0.7\degr{} in roll, and 6\% in focal length (\cref{tab:accuracy,fig:accuracy}a), and the rendered board is read almost exactly.
On identical inputs the probe is more accurate than GeoCalib in horizon (paired difference $-0.024$\ci{$-0.036$}{$-0.017$} image heights), pitch ($-0.64$\degr\ci{$-0.84$}{$-0.27$}) and, slightly, focal length ($-0.021$\ci{$-0.038$}{$-0.004$} in $|\log f|$), and less accurate in roll (\cref{sec:roll}).
The horizon advantage also holds on the original floor, on cartoon restylings and in each scene.
The pitch advantage holds on cartoons and on the original floor (supplement).
Accuracy depends on scene structure.
In the pavilion, with long architectural edges and slab joints that remain on the grey floor, the probe is nearly as exact as the oracle.
It is weakest in the cluttered classroom, where classical vanishing points on the input fail completely (supplement).
These advantages hold on renders, where clean architectural edges also make classical vanishing points on the input accurate (\cref{tab:accuracy}).
On real NYUv2 photographs the painted horizon is on par with GeoCalib (0.034 \vs 0.040 image heights), whereas its focal length (8\% \vs 4\%) and, on rotated photographs, its roll (3.0\degr{} \vs 0.6\degr) are less accurate, because the editor's priors act there (\cref{sec:prompt}); classical vanishing points on the same photographs are off by 0.209 image heights (supplement).

\noindent\textbf{Projective fidelity.}
The painted pattern is a nearly projectively correct image of a planar grid.
Its segments meet their vanishing points with a median angular residual of 0.26\degr{} (rendered board: 0.11\degr; residuals over the RANSAC inliers).
Back-projected with the true camera, the two painted families are nearly perpendicular (\cref{tab:accuracy}) and lie almost in the floor plane, and the tiles are square and correctly foreshortened, with a size that depends on the scene but not on the camera (supplement).
The editor aligns the board with the walls, so the board also measures the camera's yaw relative to the room, with a median error of 1.1\degr{} (supplement).
Painted poles converge like true verticals (parallel for shift lenses, Fig.~\ref{fig:examples}) and, with the board, give the principal point to 3.8\% of the image width.
On real photographs, however, \qwen{} paints them almost parallel, so the pole probe only works on renders.

\noindent\textbf{Not copied from the floor.}
In the classroom and the apartment, whose grey floor is featureless, removing the floor structure does not measurably reduce accuracy.
On identical cameras, the pitch error changes by $+0.13$\degr\ci{$-0.30$}{$+0.60$}, the focal error does not change significantly either, the painted tiles keep their size, and the probe still beats GeoCalib in horizon (supplement).
The editor therefore does not need floor texture and takes the painted perspective from the remaining scene structure.

\noindent\textbf{Painting versus naming.}
The same editor cannot \emph{name} the geometry it paints.
Asked to draw the horizon (\cref{fig:qualitative}e, \cref{fig:accuracy}b), \qwen{} and \kontext{} place a line at 0.49 image heights regardless of the true horizon and of camera roll, and \qwen{} does so even when the prompt explains the horizon geometrically.
Asked to mark the vanishing point, \qwen{} places its dot 0.52 image heights from the nearest true vanishing point, which is farther away than the image centre (0.45).
Camera parameters in the prompt are ignored as well.
Stating the true focal length does not change the painted perspective, requested tile sizes barely change the tiles, and a sheared board that the editor is asked to redraw with correct perspective keeps its shear (supplement).
Its geometric knowledge is therefore revealed by what it paints and not by the explicit tasks we tried.

\section{The Priors of the Implicit Camera}
\label{sec:priors}

\begin{figure*}[t]
\centering
\includegraphics[width=\textwidth]{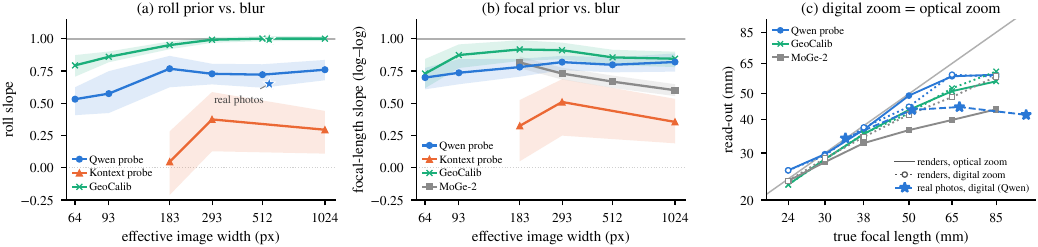}
\caption{\textbf{Priors under degraded input} (F\,$\cup$\,Z, neutral floor, one variant per image, bands show 95\% bootstrap intervals).
(a,\,b)~Roll and focal slopes stay flat down to 183\,px effective width, where four fifths of the line evidence is gone. Only at 93 and 64\,px does \qwen{} fall back further, as GeoCalib does. MoGe-2 improves with blur. Stars: rotated real photographs.
(c)~Digital zoom (crops of the 24\,mm render) is read like optical zoom, so the saturation on cropped photographs is not a cropping artefact.}
\label{fig:blur}
\end{figure*}

\begin{figure*}[t]
\centering
\includegraphics[width=\textwidth]{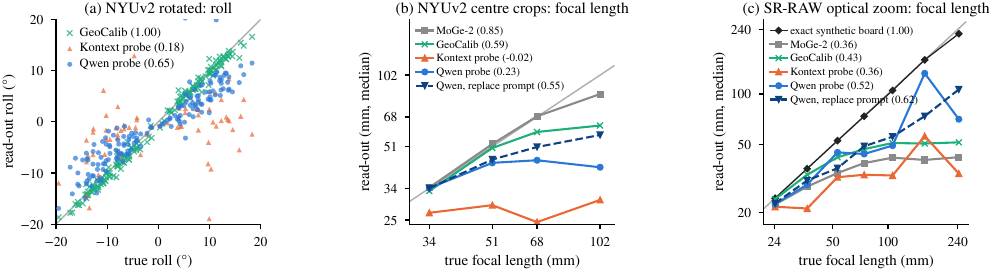}
\caption{\textbf{Real photographs} (slopes in parentheses).
(a)~NYUv2 photographs rotated by $\pm$6\degr{} and $\pm$12\degr.
(b)~NYUv2 centre crops, medians per crop level. With \pcover{} the painted focal length saturates at about 45\,mm, and \preplace{} doubles the slope.
(c)~SR-RAW optical zoom (36 sequences), medians per focal length. An exact board projected into the same frames is read nearly correctly (240\,mm as 227\,mm), whereas the painted board grows with a reduced slope and GeoCalib and MoGe-2 saturate.}
\label{fig:real}
\end{figure*}

\subsection{Roll is pulled towards level}
\label{sec:roll}
Read-out roll grows only 0.71 times as fast as true roll (\cref{fig:priors}b, \cref{tab:priors}), so a camera rolled by 12\degr{} is painted as if it were rolled by 8.5\degr.
The compression is symmetric and holds in every scene and on both floors (supplement).
Pitch, in contrast, follows 1:1, and GeoCalib follows roll exactly.

\subsection{Telephoto is pulled towards a default}
\label{sec:tele}
Up to about 40\,mm the painted focal length is exact.
For longer lenses the painted perspective stays too wide, and an 85\,mm camera is read as about 58\,mm (\cref{fig:priors}c, \cref{fig:qualitative}a,b).
The log--log fit (slope 0.81) crosses the diagonal at $f_0 = 33$\,mm (31\,mm\ci{21}{35} pooled over the blur levels down to 183\,px), so long focal lengths are pulled towards a default of about 30\,mm.
Scene structure modulates the pull: in the pavilion, with its long architectural edges, the focal slope is 0.97 (supplement).
The read-out does not cause it, because the rendered board follows the lens.

This default roughly matches the camera the models paint when there is no scene to follow (Fig.~\ref{fig:default}).
Asked to turn a blank grey image into a room with a checkerboard floor, \qwen{} paints a level, slightly downward camera with $f = 32$\,mm (median of 10 seeds, measured with GeoCalib and hence indicative).
All other models we tested (\kontext, \longcat, OmniGen2, \qwen's text-to-image mode and Krea~2 Turbo) paint an unrolled camera between 23 and 36\,mm, with pitch varying between models (supplement).
Lens instructions hardly move the default (``85mm telephoto lens'': 32\,mm with \qwen), and a requested 15\degr{} Dutch angle arrives as 0.7\degr{} of roll.

\subsection{The priors do not grow with line blur}
\label{sec:blur}
If the priors were a precision-weighted compromise between image evidence and a learned prior, they should grow as the evidence fades.
To test this, we blur the inputs at a fixed camera (\cref{fig:blur}).
Down to 183\,px effective width, where four fifths of the floor-line evidence usable by classical methods is gone, both slopes stay flat (all paired changes within $\pm$0.04 and not significant) and pitch stays exact (supplement).
Only when even coarse structure disappears (significantly at 64\,px) does \qwen{} fall back further, towards the same level default (roll slope 0.53 at 64\,px, paired change $-0.23$\ci{$-0.36$}{$-0.10$}), as GeoCalib does.
For \qwen, the implicit camera thus combines a bias that is present at every resolution with a fallback to its default when coarse structure vanishes. This need not hold for other editors: \kontext's roll slope falls from 0.30 to 0.05 at 183\,px, although not significantly (22 paired renders, supplement). The line evidence the editor itself uses is not measured.

\subsection{Real photographs and the wording of the task}
\label{sec:prompt}
On real photographs both priors are stronger (\cref{fig:real}, \cref{tab:priors}), although the painted board still recovers the horizon (\cref{fig:qualitative}d).
On rotated NYUv2 photographs the roll slope is 0.65, and on centre crops the painted focal length saturates at about 45\,mm (slope 0.23).
GeoCalib follows roll exactly on the same photographs.
The manipulations are not the cause.
The same rotation and cropping applied to renders gives a roll slope of 0.85, NYU-like image statistics lower it only to 0.78, digital zoom is read like optical zoom (\cref{fig:blur}c), and blurring the photographs leaves their roll slope unchanged (supplement).

On NYUv2, the roll gap depends on the wording of the task (\cref{tab:priors}, lower block).
With \preplace{} the roll slope on the same photographs rises from 0.63 to 0.78 (paired $+0.15$\ci{$+0.07$}{$+0.20$}), which is the value this prompt reaches on renders, and the focal slope doubles.
Rotated SR-RAW frames, which show outdoor scenes at higher resolution, point the same way without confirming it.
Over all variants, as specified before the runs, the roll slope rises by only $+0.09$\ci{$-0.01$}{$+0.20$} over sequences, in the first variant from 0.65 to 0.83 ($+0.18$\ci{$+0.07$}{$+0.30$}), while GeoCalib follows exactly (supplement).
On renders the short prompt leaves roll unchanged, improves the focal slope and more than halves the horizon error (supplement).
The roll prior is therefore a property of the model on renders and photographs alike, and on NYUv2 the longer instruction amplifies it (and the focal prior).
We recommend \preplace{} as the probe.

\noindent\textbf{Optical zoom.}
With real lenses the telephoto prior persists (\cref{fig:real}c, \cref{tab:priors}).
On the SR-RAW frames the painted focal length grows with slope 0.62 under \preplace{} (0.52 under \pcover, which does not differ significantly), and the 240\,mm frames are painted like 106\,mm ones.
An exact board projected into the same frames, however, is read correctly (slope 1.00).
Optical and digital zoom to the same field of view are read alike, so cropping does not cause the saturation on NYUv2 (supplement).
Learned calibrators saturate as well and do so earlier, with GeoCalib reading the 240\,mm frames as 52\,mm and MoGe-2 as 42\,mm.
On identical frames the probe follows the lens as far as GeoCalib (paired slope difference $+0.13$\ci{$-0.01$}{$+0.28$}) and further than MoGe-2 ($+0.24$\ci{$+0.13$}{$+0.37$}).
On real photographs, a telephoto deficit is therefore not specific to the editor, at least among the models we tested.

\subsection{Editors differ}
\label{sec:editors}
\kontext{} and \longcat{} are pulled much harder (\cref{tab:priors}, \cref{fig:qualitative}c).
On renders their roll slopes are 0.30 and 0.25 and their focal slopes 0.43 and 0.34, with fixed points of 25 and 28\,mm.
On NYUv2 crops, \kontext{} paints a nearly constant perspective of 24--30\,mm regardless of the crop.
Both paint less rectangular grids and are less accurate than \qwen{} on every measure (\cref{tab:accuracy}), and only half of \longcat's outputs yield a readable board with a valid registration (supplement).
The two priors appear in all three editors, and their strength separates \qwen{} from the other two.

\section{Writing versus Reading the Camera}
\label{sec:writeread}

\begin{figure}[t]
\centering
\includegraphics[width=\spalte]{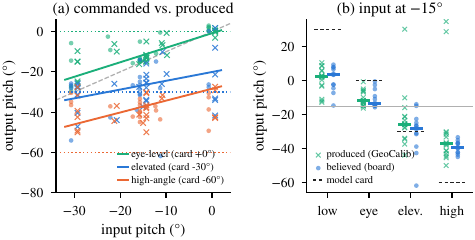}
\caption{\textbf{Writing \vs reading the camera} with a multi-angle LoRA.
(a)~Output pitch against input pitch for three commands. Crosses give the camera produced (GeoCalib on the output), dots the camera read from a board that the editor, without the LoRA, paints into the output, and dotted lines the model-card targets.
(b)~For inputs at $-15$\degr, commands are executed at reduced strength, but the painted board agrees with the produced camera.}
\label{fig:lora}
\end{figure}

Camera-control adapters write a camera, and our probe can read it.
We apply a multi-angle LoRA for Qwen-Image-Edit-2511~\cite{fal2026multipleangles} with seven commands to 16 catalog renders, and with four of them to 12 further renders at 0\degr{} and $-30$\degr{} pitch (supplement).
GeoCalib on the output gives the camera the model produced. In a second call without the LoRA, the editor paints our board into the output, which gives the camera the editor assumes for the generated view.

\noindent\textbf{Commands are executed at reduced strength} (\cref{fig:lora}).
From a level input, where the commands are defined, ``elevated shot'' (30\degr{} per model card) tilts the camera to $-20$\degr{} and ``high-angle shot'' (60\degr) to $-28$\degr.
The response is neither absolute nor relative to the input pitch (\cref{fig:lora}a and supplement).
A ``front-right quarter view'' (45\degr) turns the camera by only 30\degr{}, measured with the yaw read from the board, while the eye-level control does not turn it (supplement).

\noindent\textbf{The painted board agrees with the produced camera.}
For every command, the pitch read from the board differs from GeoCalib's by 1.4--3.4\degr{} (median per command), comparable to the 2.0\degr{} on the unedited renders.
The editor therefore reads generated views as consistently as renders, even when the produced camera is not the commanded one, and the probe can verify camera commands without ground truth.
The probe also tracks generated videos frame by frame.
When LTX-2.5~\cite{lightricks2026ltx25,hacohen2026ltx2} interpolates between two renders with a physical board, it moves the camera monotonically and keeps the focal length within 3\% of the true value (supplement).

\section{Discussion and Limitations}
\label{sec:discussion}

\noindent\textbf{Where do the priors come from?}
Both priors point to the same camera, a level one at about 30\,mm.
As far as GeoCalib can measure it, five different models paint this camera when given no scene at all (\cref{sec:tele}).
As the priors do not grow with uncertainty (\cref{sec:blur}), the editor seems to infer its camera from the coarse structure of the scene and bias it towards this default, more or less strongly depending on the wording (\cref{sec:prompt}).
That roll is compressed while pitch is not is consistent with training images that are mostly level but taken from many heights and angles.
Testing this would require access to the training data.
On optically zoomed photographs GeoCalib and MoGe-2 saturate as well, although our read-out recovers long focal lengths from an exact board projected into the same frames.
Underestimating long focal lengths therefore seems to be common to the models we tested that learn the camera from images and is not specific to editors.

\noindent\textbf{Implications.}
(i)~Editors used for compositing or data synthesis are likely to inherit a level camera of about 30\,mm.
People they insert into 85\,mm views are painted too small (for adults of 1.72\,m, the implied camera is 2.2--3.8$\times$ too high), more than the board's telephoto prior explains (supplement).
(ii)~The implicit camera is a measurable model property that separates \qwen{} from the other editors and could serve as a benchmark axis.
(iii)~To read geometric knowledge out of a generative model, ask it to paint structure rather than to name geometry, and treat the wording of the task as part of the measurement.

\noindent\textbf{Limitations.}
Three open editors (\longcat{} on renders only) and three indoor scenes are a small sample of a fast-moving field, although the protocol applies to any editor that accepts an image and a prompt.
We measured OmniGen2 only for the default camera.
Our real photographs are NYUv2 indoor scenes and SR-RAW sequences, 35 of 36 of them outdoors, with synthetically varied cameras or optical zoom.
On the zoom frames the probe yields a focal length for only 61\% of the frames, and its slope depends on the vanishing-point bound, although it stays below 1 under every bound (supplement).
The read-out assumes square pixels, a centred principal point unless poles are painted, and a visible floor.
Poles work on renders but not on real photographs.
Several controls use a single variant per image, and the control against copying rests on two scenes, because the pavilion's grey floor keeps its slab joints.
All \qwen{} results use the 4-step Lightning distillation, compared with the full model in one scene only.
Paired comparisons with GeoCalib include only images with a readable board (88\% of the catalog for pitch and focal length) and use the median over three editor seeds against a single GeoCalib run.
Readability can also select paintings: the vanishing-point bound decides which frames yield a focal length (supplement).
Finally, we measure the camera an editor encodes when painting this pattern.
Inserted people shrink in the same direction but more strongly, so other edits share at least part of this bias, but their camera is not fully explained by the board's.

\section{Conclusion}
\label{sec:conclusion}
Asking an image editor to paint a checkerboard turns it into a calibration target and makes the camera it assumes measurable.
This implicit camera is precise on renders, but it is pulled towards a level default of roughly 30\,mm, more strongly on real photographs and depending on how the task is phrased.
Editors also paint geometry that they cannot name.
We hope that painted probes become a routine instrument for measuring what generative models assume about the camera.

\ifdefined\ReviewVersion
\section*{Disclosure}
An AI agent (Claude, Anthropic) was used in the implementation, analysis and preparation of this work and in the language editing of the manuscript.
All results were checked by the authors, who take full responsibility for the content.
\else
\section*{Disclosure}
An AI agent (Claude, Anthropic) was used in the implementation, analysis and preparation of this work and in the language editing of the manuscript.
All results were checked by the author, who takes full responsibility for the content.
\fi

\bibliographystyle{tmlr}
\bibliography{literatur}

\supplementanfang 
\appendix
\setcounter{figure}{0}\setcounter{table}{0}
\renewcommand{\thefigure}{S\arabic{figure}}\renewcommand{\theHfigure}{S\arabic{figure}}
\renewcommand{\thetable}{S\arabic{table}}\renewcommand{\theHtable}{S\arabic{table}}
\setlength{\textfloatsep}{10pt plus 2pt minus 2pt}\setlength{\dbltextfloatsep}{10pt plus 2pt minus 2pt}
\setlength{\floatsep}{8pt plus 2pt minus 2pt}\setlength{\dblfloatsep}{8pt plus 2pt minus 2pt}
\setlength{\abovecaptionskip}{4pt}

\section{Prompts}
\label{sec:suppl_prompts}
All prompts were used verbatim, and the editors saw no other text.

\noindent\textbf{Probes.}
\pcover: \emph{``Cover the entire visible floor with a large black and white checkerboard tile pattern made of square tiles, lying perfectly flat on the floor and following the perspective of the floor. Furniture stays on top of the floor. Keep everything else in the image exactly the same.''}
\preplace: \emph{``Replace the floor with a black-and-white checkerboard floor.''}
Paraphrase: \emph{``Lay a chessboard-like floor of alternating dark and light square tiles over the whole floor, in correct perspective. Do not change anything else.''}
Poles: \emph{``Add six tall, thin, perfectly straight vertical poles in solid pure magenta (\#FF00FF), standing upright on the floor at clearly different places and distances, each reaching from the floor up to the ceiling or out of the top of the picture. The poles are exactly vertical like plumb lines and do not overlap each other. Keep everything else in the image exactly the same.''}

\noindent\textbf{Explicit tasks.}
Horizon: \emph{``Draw the horizon line of this photo: one thin, perfectly straight, solid pure magenta (\#FF00FF) line across the entire width of the image, exactly at the eye level of the camera, where the floor would vanish at infinity if the room had no walls. Draw it on top of everything. Keep everything else in the image exactly the same.''}
Horizon, explained: \emph{``Draw one thin, perfectly straight, solid pure magenta (\#FF00FF) line across the whole image exactly at the camera's eye level: the horizon line, where all parallel horizontal lines of the floor would meet if they were extended. Keep everything else in the image exactly the same.''}
Vanishing point: \emph{``Mark the vanishing point of this photo: draw one small solid pure magenta (\#FF00FF) dot exactly at the point where the parallel lines of the floor and the walls converge. Keep everything else in the image exactly the same.''}

\noindent\textbf{Camera parameters in the text.}
Lens: \emph{``This photo was taken with a \{16, \ldots, 85\} mm lens (full-frame equivalent). Cover the entire visible floor with a large black and white checkerboard tile pattern made of square tiles, lying perfectly flat on the floor and following the perspective of the floor and of this lens. Furniture stays on top of the floor. Keep everything else in the image exactly the same.''}
Tile size: \emph{``Cover the entire visible floor with a black and white checkerboard pattern of square floor tiles, each tile exactly 50 cm by 50 cm, lying perfectly flat on the floor and following the perspective of the floor. Furniture stays on top of the floor. Keep everything else in the image exactly the same.''} and the same with \emph{``large square floor tiles, each tile exactly 1 m by 1 m''}.
Correction: \emph{``The black and white checkerboard pattern on the floor has a wrong perspective. Redraw it as a regular checkerboard of square tiles that lies perfectly flat on the floor with exactly the correct perspective of this photo. Keep everything else in the image exactly the same.''}
Replacement: \emph{``Replace the pattern on the floor with a new regular black and white checkerboard of square tiles that lies perfectly flat on the floor and follows the perspective of the floor. Furniture stays on top of the floor. Keep everything else in the image exactly the same.''}

\noindent\textbf{Default camera.}
Editor on a blank grey $1024\times768$ image: \emph{``Turn this empty image into a photo of a large empty room whose whole floor is covered with a large black and white checkerboard tile pattern made of square tiles.''}
Text-to-image: \emph{``A photo of a large empty room. The whole floor is covered with a large black and white checkerboard tile pattern made of square tiles.''}, optionally followed by \emph{``Shot with a 24mm wide-angle lens.''}, \emph{``Shot with an 85mm telephoto lens.''} or \emph{``Dutch angle: the camera is rolled 15 degrees to the side.''}

\noindent\textbf{Multi-angle LoRA.}
The LoRA supports 8 azimuths, 4 elevations and 3 distances.
We use commands in the format of its model card~\cite{fal2026multipleangles}: \texttt{<sks> front view \{eye-level, elevated, high-angle, low-angle\} shot medium shot}, \texttt{<sks> front view eye-level shot \{close-up, wide shot\}} and \texttt{<sks> front-right quarter view eye-level shot medium shot}.
The painted board was added to the LoRA output with \pcover.

\section{Read-out Details and Robustness}
\label{sec:suppl_readout}

\noindent\textbf{Registration and masks.}
ORB uses 4000 features on grey images with the probe colours masked out.
The similarity transform is estimated by RANSAC with a 2\,px reprojection threshold.
On the catalog, \qwen{} rescales its output by 0.997--1.002 and shifts it by at most 0.54\% of the image height, \kontext{} by 0.993--1.000 (median 0.999) and 0.82\%, and \longcat{} by 0.958--1.013 (median 0.996) and 2.5\% (registrations with at least 20 inliers).
Registrations supported by fewer than 20 inliers that deviate from the identity by more than 2\% in scale or shift or by more than 1\degr{} in rotation are treated as failures, because RANSAC then returns scales between 0.003 and 3.
Such variants are discarded, which affects 2\% of the \qwen{} variants on the neutral floor, 11\% on the original floor, 20\% of the \kontext{} and 44\% of the \longcat{} variants.
On real photographs the probe is read in the edited image itself, where the editor's drift is negligible.
The change mask thresholds the Gaussian-blurred ($\sigma = 3$\,px) absolute grey-level difference at 25 and the blurred ($\sigma = 5$\,px) Laplacian magnitude of the edited image at its median, followed by morphological closing and opening.
The painted region covers 18\% of a sharp input and 19--23\% of the blurred inputs, \ie the editor does not re-sharpen the scene.

\noindent\textbf{Robustness to the thresholds.}
\cref{tab:suppl_robust} re-evaluates the neutral-floor catalog with seven settings of the three thresholds of the floor-pair selection.
Slopes and the pitch error differ from the default by at most 0.03 and 0.09\degr.

\begin{table}[h]
\centering
\caption{Robustness of the read-out to its thresholds (neutral floor, sets F\,$\cup$\,Z, 119 evaluable images): fraction of a pair's segments below its vanishing line, minimum direction difference of the two families, and RANSAC inlier angle. First row: default.}
\label{tab:suppl_robust}
\footnotesize
\setlength{\tabcolsep}{3.6pt}
\begin{tabular}{@{}cccccc@{}}
\toprule
below line & direction & RANSAC & roll slope & $f$ slope & pitch err. \\
\midrule
0.9 & 10\degr & 1.5\degr & 0.71 & 0.81 & 0.76\degr \\
0.8 & 10\degr & 1.5\degr & 0.71 & 0.81 & 0.76\degr \\
0.95 & 10\degr & 1.5\degr & 0.71 & 0.81 & 0.76\degr \\
0.9 & 5\degr & 1.5\degr & 0.71 & 0.81 & 0.76\degr \\
0.9 & 15\degr & 1.5\degr & 0.71 & 0.81 & 0.76\degr \\
0.9 & 10\degr & 1.0\degr & 0.68 & 0.82 & 0.67\degr \\
0.9 & 10\degr & 2.5\degr & 0.71 & 0.81 & 0.75\degr \\
\bottomrule
\end{tabular}
\end{table}

\noindent\textbf{Line evidence.}
To quantify how much perspective information an input offers to a line-based method, we sum the lengths of all LSD segments of the input that point to a true floor vanishing point within 1.5\degr, in units of the image diagonal.
Among the sharp images with $f > 45$\,mm, more evidence goes with less focal compression by the editor (Spearman $\rho = 0.43$, $p = 0.03$, $n = 26$).
Across scenes and blur levels, the roll slopes of \qwen{} and GeoCalib follow the evidence ($\rho = 0.62$ and 0.91 over 18 scene--level cells).

\section{Catalog and Controls}
\label{sec:suppl_catalog}

\noindent\textbf{Scenes.}
Classroom by Christophe Seux (CC0), Barcelona Pavillion by eMirage/Hamza Cheggour after a model by Claudio Andres (CC-BY), and Italian Flat by Flavio Della Tommasa (CC-BY), from the Blender demo files.
We added our own cameras, floors and boards and rendered new images.
The classroom includes an empty side room, so some of its random cameras show no furniture.

\noindent\textbf{Cameras.}
All cameras have a 36\,mm wide sensor and no depth of field.
Set F uses the lenses 16, 20, 24, 30, 38, 50, 65 and 85\,mm at two positions per scene.
Set Z draws the focal length log-uniformly, the camera height from 0.9 to 2.0\,m and the yaw freely, and accepts a camera only if ray casting finds free space in front of it and at least 20\% of the image shows floor.
Set S uses level cameras with a vertical lens shift of $-0.12$ or $-0.2$ (in units of the sensor width), which moves the principal point to 0.34 or 0.23 image heights from the top instead of 0.5.
Ground truth consists of $K$, $R$ and the camera centre.
The horizon and the vanishing points of the room axes follow from them and agree with Blender's own projection to 0.001\,px.

\noindent\textbf{Controls.}
The neutral floor replaces the floor material by a uniform grey.
In the pavilion, the slab joints are part of the geometry and remain visible. The oracle renders a 0.5\,m checkerboard as floor texture.
The blur ladder downsamples by $r$ and upsamples back to $1024\times768$, both with Lanczos filtering (effective widths 512 to 64\,px).
Digital zoom crops the centre of the 24\,mm render of set F to the field of view of 30, 38, 50, 65 and 85\,mm and upsamples it, exactly as for the photographs.
For rotation we rotate the image by $\pm$6\degr{} or $\pm$12\degr{} about its centre (bicubic), crop the largest centred rectangle without black corners, upsample it (Lanczos) and update the camera exactly.
The photo look additionally applies downsampling to $552\times414$, desaturation, a warm colour cast, a gamma change, Gaussian noise ($\sigma = 5$ grey levels) and JPEG compression at quality 50.
The editor restyles the 48 images of set~F as cartoons.

\noindent\textbf{Real photographs.}
Our NYUv2 set consists of 40 test images in which the floor covers at least 8\% of the image.
The digital zoom crops them by factors of 1.5, 2 and 3 and upsamples the crops to $640\times480$, which corresponds to 51, 68 and 102\,mm instead of 34\,mm.
Rotation and zoom give 160 images each.
For the rotated SR-RAW frames we rotate the widest optical frame of each of the 36 sequences by $-12$, $-6$, 0, $+6$ and $+12$\degr{} and crop all five by the same factor of 1.29, the smallest one that leaves no black corners at 12\degr{}, so that only the roll differs between them (180 frames at $1152\times768$, 31\,mm equivalent).
The true roll of these outdoor frames is unknown, so we evaluate the change of roll against the unrotated frame of the same sequence, and resample sequences for the confidence intervals.

\section{Additional Results}
\label{sec:suppl_results}

\begin{figure*}[t]
\centering
\includegraphics[width=0.85\textwidth]{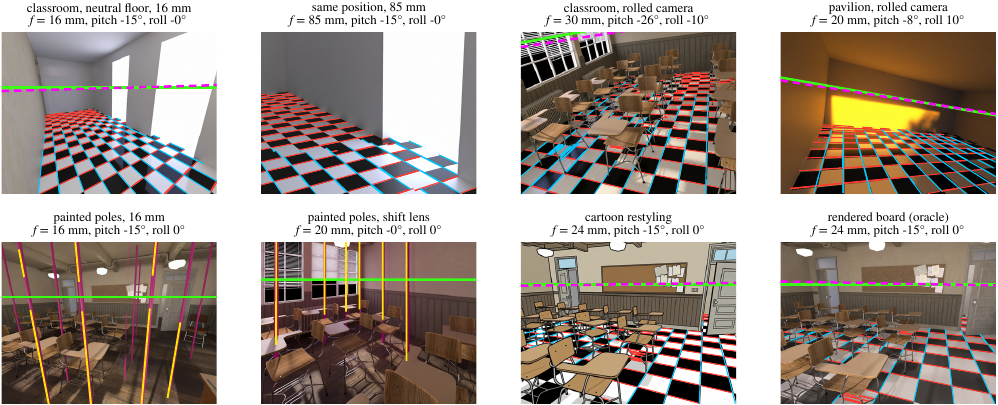}
\caption{\textbf{Catalog examples} (\qwen, one variant each). Titles give the true camera. Green: true horizon, dashed magenta: horizon read from the painted board, red/blue: detected floor families, yellow: pole axes. Top: the classroom at 16\,mm and at 85\,mm from the same position (telephoto painted too wide), and rolled cameras in the classroom and the pavilion, read as $-8.9$\degr{} for $-10.0$\degr{} and 8.9\degr{} for 10.0\degr{} (roll compressed). Bottom: painted poles in three-point perspective at 16\,mm and parallel for a shift lens, a cartoon restyling and the rendered oracle board.}
\label{fig:examples}
\end{figure*}

\begin{figure*}[t]
\centering
\includegraphics[width=0.85\textwidth]{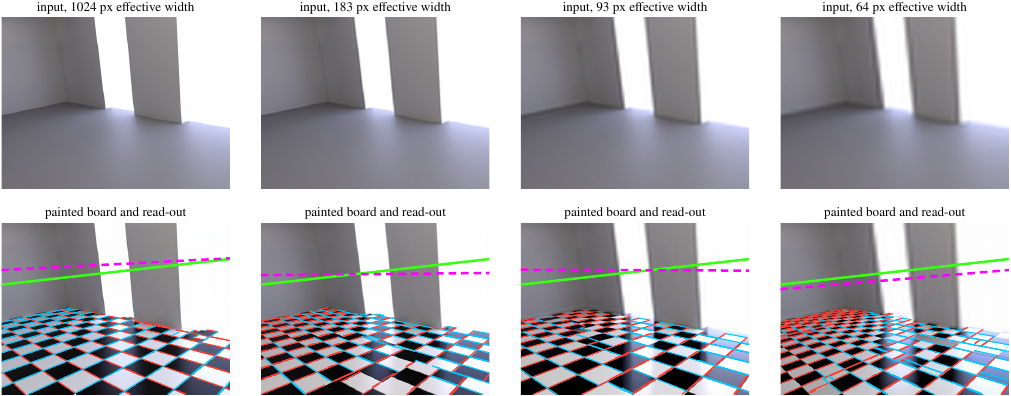}
\caption{\textbf{Blur ladder examples.} Input at 1024, 183, 93 and 64\,px effective width (top) and the board painted by \qwen{} (bottom), with the read-out as in \cref{fig:examples}. The editor paints a sharp board into a blurred scene without re-sharpening the scene.}
\label{fig:suppl_blur}
\end{figure*}

\begin{table}[t]
\centering
\caption{\textbf{Blur ladder} (sets F\,$\cup$\,Z, neutral floor, 120 renders per level, one variant): effective width, line evidence (image diagonals), \qwen's roll, focal and pitch slopes and horizon error, and GeoCalib's (GC) slopes. Paired against the sharp level, \qwen's roll slope changes by $-0.03$, $-0.03$ and $+0.01$ up to $r = 5.6$ (all intervals cover zero), and by $-0.16$\ci{$-0.31$}{$+0.03$} and $-0.23$\ci{$-0.36$}{$-0.10$} at $r = 11$ and 16. The fixed point of the focal fit is 31\,mm\ci{21}{35} pooled up to $r = 5.6$, and 27 and 28\,mm at $r = 11$ and 16. \kontext{} (80 renders): roll 0.30, 0.38, 0.05 and focal 0.36, 0.51, 0.32 at $r = 1$, 3.5, 5.6.}
\label{tab:suppl_blur}
\footnotesize
\setlength{\tabcolsep}{2.4pt}
\begin{tabular}{@{}rrrcccccc@{}}
\toprule
$r$ & px & evid. & roll & $f$ & pitch & hor. & GC roll & GC $f$ \\
\midrule
1 & 1024 & 6.0 & 0.76 & 0.82 & 0.99 & 0.028 & 1.00 & 0.85 \\
2 & 512 & 4.7 & 0.72 & 0.80 & 1.00 & 0.024 & 1.00 & 0.86 \\
3.5 & 293 & 2.7 & 0.73 & 0.82 & 1.05 & 0.028 & 0.99 & 0.91 \\
5.6 & 183 & 1.1 & 0.77 & 0.78 & 1.03 & 0.044 & 0.95 & 0.92 \\
11 & 93 & 0.2 & 0.58 & 0.74 & 1.00 & 0.046 & 0.86 & 0.87 \\
16 & 64 & 0.0 & 0.53 & 0.70 & 1.08 & 0.042 & 0.79 & 0.73 \\
\bottomrule
\end{tabular}
\end{table}

\noindent\textbf{Blur ladder.}
\cref{tab:suppl_blur} lists all levels, and \cref{fig:suppl_blur} shows an example.
At every level \qwen{} remains evaluable on 88--98\% of the images, GeoCalib and MoGe-2 on all, and \kontext{} on 72--75\%.
MoGe-2 estimates the focal length better the blurrier its input (slope 0.60 at full resolution, 0.82 at 183\,px).
The editor's errors are properties of the image, because they are reproducible between the sharp and the 183\,px version (Spearman $\rho = 0.60$ for roll and 0.67 for focal length).
Its roll errors are uncorrelated with GeoCalib's errors on the same sharp images ($\rho = -0.04$), whereas focal errors are partly shared ($\rho = 0.37$).
Blurring the rotated real photographs to 276 and 158\,px gives roll slopes of 0.65 and 0.60, against 0.66 for the sharp images (paired differences $+0.01$\ci{$-0.07$}{$+0.08$} and $-0.05$\ci{$-0.13$}{$+0.03$}).

\noindent\textbf{Controls for real photographs and the wording.}
Rotating and cropping renders like the photographs gives a roll slope of 0.85\ci{0.81}{0.90}, and adding the NYU-like photo look lowers it only to 0.78 (paired $-0.08$\ci{$-0.13$}{$-0.03$}).
Digital zoom of renders is read like optical zoom (focal slope 0.75 \vs 0.77).
With one variant per image on set~Z, \preplace{} leaves the roll slope unchanged (0.76 \vs 0.78, a paraphrase gives 0.75), raises the focal slope from 0.76 to 0.88 and lowers the horizon error from 0.041 to 0.015 image heights (\cref{tab:priors}).
For \kontext{} it does not help (roll slope 0.34).

\begin{figure*}[t]
\centering
\includegraphics[width=0.85\textwidth]{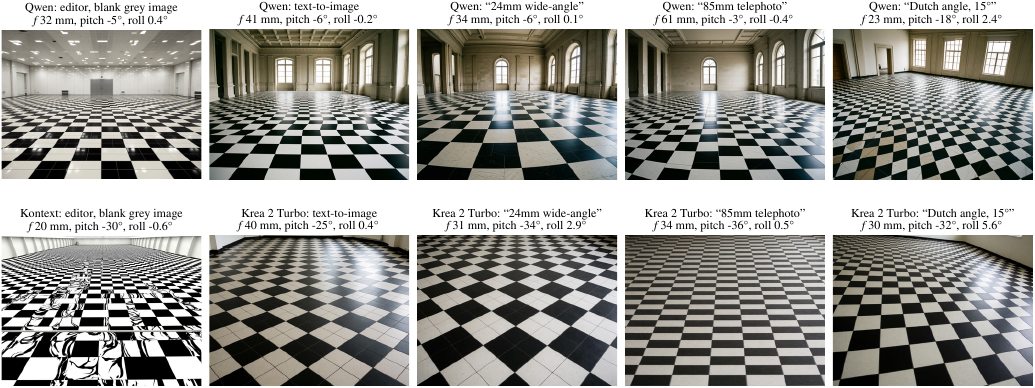}
\caption{\textbf{The default camera}: rooms generated without any scene to follow (one seed each, GeoCalib read-out above each image). All models paint a level camera with a moderate wide-angle focal length. Lens instructions barely change the focal length, and a requested 15\degr{} Dutch angle barely changes the roll.}
\label{fig:default}
\end{figure*}

\noindent\textbf{Default camera.}
\cref{fig:default} shows one seed per condition and \cref{tab:suppl_default} the medians over 10 seeds, measured with GeoCalib on the generated images.
The values are indicative, because GeoCalib itself pulls focal lengths towards 20\,mm (slope 0.85 on the catalog).
They agree with the fixed point of 33\,mm measured with the board.

\begin{table}[t]
\centering
\caption{\textbf{Default camera without scene cues}: medians over 10 seeds (GeoCalib on the generated images).}
\label{tab:suppl_default}
\footnotesize
\setlength{\tabcolsep}{3.2pt}
\begin{tabular}{@{}llccc@{}}
\toprule
Model & Condition & $f$ (mm) & Pitch & Roll \\
\midrule
\qwen{} editor & blank grey image & 32 & $-5.2$\degr & 0.1\degr \\
\kontext{} editor & blank grey image & 26 & $-17.2$\degr & $-0.5$\degr \\
\longcat{} editor & blank grey image & 23 & $-9.6$\degr & $-0.1$\degr \\
OmniGen2 editor & blank grey image & 29 & $-9.7$\degr & 0.2\degr \\
\qwen{} text-to-im. & neutral & 31.5 & $-7.0$\degr & $-0.1$\degr \\
 & ``24mm wide-angle'' & 25.6 & $-8.9$\degr & 0.3\degr \\
 & ``85mm telephoto'' & 32.2 & $-5.8$\degr & $-0.1$\degr \\
 & ``Dutch angle, 15\degr'' & 19.2 & $-19.4$\degr & 0.7\degr \\
Krea 2 Turbo & neutral & 35.8 & $-26.9$\degr & 0.1\degr \\
 & ``24mm wide-angle'' & 30.6 & $-32.8$\degr & $-0.3$\degr \\
 & ``85mm telephoto'' & 39.5 & $-35.8$\degr & 1.8\degr \\
 & ``Dutch angle, 15\degr'' & 31.7 & $-32.1$\degr & 0.9\degr \\
\bottomrule
\end{tabular}
\end{table}

\noindent\textbf{Projective fidelity per scene.}
The median angular residual of the painted segments to their vanishing point is 0.32\degr{} (classroom), 0.24\degr{} (pavilion) and 0.25\degr{} (apartment) for \qwen{} on the neutral floor (original floor: 0.35\degr, 0.32\degr{} and 0.24\degr), and 0.09\degr, 0.13\degr{} and 0.10\degr{} for the rendered board.
Back-projected with the true camera, the painted families are tilted by 1.1\degr{} against the floor plane (median).
The editor aligns the board with the room, with 92\% of the images within 5\degr{} of the wall directions, and the rendered board gives the yaw to 0.1\degr.
Per scene on the neutral floor, the \qwen{} probe reaches a horizon error of 0.044, 0.006 and 0.027 image heights, a pitch error of 1.1\degr, 0.2\degr{} and 1.2\degr, and a focal slope of 0.61, 0.97 and 0.80 (classroom, pavilion, apartment).

\noindent\textbf{Accuracy against GeoCalib by floor, style and scene} (paired differences probe $-$ GeoCalib on identical images).
Over all scenes on the neutral floor, the focal error differs by $-0.021$\ci{$-0.038$}{$-0.004$} and the roll error by $+0.27$\degr\ci{$+0.08$}{$+0.67$}.
Per scene on the neutral floor, the horizon error is lower in all three scenes (classroom $-0.026$\ci{$-0.057$}{$-0.012$}, pavilion $-0.016$\ci{$-0.035$}{$-0.007$}, apartment $-0.026$\ci{$-0.043$}{$-0.019$}) and the pitch error in the classroom ($-0.68$\degr\ci{$-1.22$}{$-0.15$}) and the pavilion ($-1.10$\degr\ci{$-1.48$}{$-0.71$}).
In the apartment it is equal ($-0.08$\degr\ci{$-0.46$}{$+0.32$}).
On the original floor, which slightly helps GeoCalib (its horizon error is 0.003\ci{0.000}{0.006} lower than on the neutral floor), the probe is more accurate in horizon ($-0.021$\ci{$-0.027$}{$-0.013$}) and pitch ($-0.33$\degr\ci{$-0.51$}{$-0.10$}), on par in focal length ($-0.017$\ci{$-0.032$}{$-0.000$}), and less accurate in roll ($+0.47$\degr\ci{$+0.22$}{$+0.74$}).
Removing the floor structure (grey \vs original floor, classroom and apartment) changes the probe's focal error by $+0.017$\ci{$-0.007$}{$+0.040$} on identical cameras.
On the 48 cartoon restylings of set~F it is more accurate in horizon ($-0.024$\ci{$-0.036$}{$-0.013$}) and pitch ($-0.46$\degr\ci{$-0.88$}{$-0.21$}) and on par in focal length ($-0.014$\ci{$-0.071$}{$+0.022$}).

\noindent\textbf{Roll prior per scene.}
The roll slope on set~Z is 0.71, 0.84 and 0.60 on the neutral floor and 0.66, 0.80 and 0.76 on the original floor (classroom, pavilion, apartment), all with confidence intervals below 1.
The compression is symmetric (intercept $+0.03$\degr).
The compression tends to be stronger for steep downward views (0.70\ci{0.60}{0.78} below $-15$\degr{} pitch \vs 0.74\ci{0.62}{0.87} above, with overlapping intervals).

\noindent\textbf{Painted tiles.}
Without a size in the prompt the editor paints tiles of 0.30--0.31\,m in the classroom, 0.51\,m in the pavilion and 0.25\,m in the apartment (medians on the original and the neutral floor), uncorrelated with camera height and focal length ($|r| \le 0.19$).
In the pavilion, the tiles follow the slab joints, which remain on the grey floor.
Asking for ``50\,cm by 50\,cm'' or ``1\,m by 1\,m'' tiles leaves the classroom tiles at 0.29 and 0.37\,m.
In the pavilion, whose slabs measure 0.5 by 1\,m, the requested 1\,m tiles are met (1.01\,m).
The painted tiles are square (aspect 0.99) and correctly foreshortened (far/near size ratio 0.99), and the rendered board is read at its true tile size (0.50\,m).
As a metric ruler for the camera height the painted tiles are therefore unreliable.
A sheared board (64\degr{} instead of 90\degr) stays at 65.4\degr{} when the editor is asked to redraw it with correct perspective, and at 65.3\degr{} when asked to replace it.

\noindent\textbf{Explicit tasks.}
The line drawn when the editor is asked for the horizon has Theil--Sen slopes between $-0.03$ and 0.00 against the true horizon height and against the true roll (\qwen, \qwen{} with the explained prompt and \kontext, with 47--72 variants each).
Stating the true focal length in the prompt changes the focal error by $-0.003$\ci{$-0.014$}{$+0.001$} (paired on identical images).

\noindent\textbf{Poles.}
On the 132 renders with the original floor (including set~S), the painted poles are 3.2\degr{} from the plumb line, and poles and board give the principal point to 3.8\% of the image width and 5.2\% of its height and the focal length to 7\%.
On digitally zoomed NYUv2 photographs \qwen{} paints the poles nearly parallel.
The focal length from horizon and zenith then grows from 55 to 323\,mm for true focal lengths of 34 to 68\,mm, whereas the board alone gives 34--46\,mm.
Poles improve the telephoto focal length on renders (original floor, set~F at 50--85\,mm: $|\log f|$ error 0.105 from horizon and zenith \vs 0.199 from the board alone) but not on photographs.

\noindent\textbf{Accuracy on NYUv2.}
On the 40 unmodified photographs the painted horizon is off by 0.034 image heights, GeoCalib's by 0.040 and classical vanishing points by 0.209.
The focal length is off by 8\% (GeoCalib: 4\%).

\noindent\textbf{NYUv2 horizon offset.}
Against the depth-derived horizon of NYUv2, the two line-based methods share an offset (GeoCalib $+0.034$, our probe $+0.030$ image heights), whereas depth-based MoGe-2 with the oracle floor mask has none ($-0.001$).
Undistorting the images with the NYUv2 calibration does not change this offset, and it cancels in all our comparisons, which use the photograph's own ground truth.

\noindent\textbf{Combining the probe with GeoCalib.}
The probe yields no pitch for 12\% of the catalog images on the neutral floor and 16\% on the original floor (no admissible floor pair, or an undefined focal length).
With GeoCalib as a fallback for these images, the median pitch error over all 120 cameras is 0.77\degr{} and 0.94\degr{} (GeoCalib alone: 1.60\degr{} and 1.47\degr). The 90th percentiles are 3.2\degr{} and 2.9\degr{} \vs 4.3\degr{} and 3.2\degr.
Using the probe only when it agrees with GeoCalib within 3\degr{} or 5\degr{} brings no further gain.
Dividing the probe's roll by the slope measured on the catalog (0.71) reduces the roll error on rotated photographs from 3.0\degr{} to 2.3\degr{} (GeoCalib: 0.6\degr).

\begin{figure}[t]
\centering
\includegraphics[width=0.85\spalte]{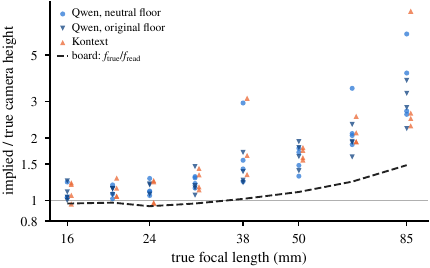}
\caption{\textbf{Inserted people shrink at long focal lengths.} Camera height implied by painted photorealistic people (assuming 1.72\,m adults), relative to the true height, on set F. Dashed: focal compression of the painted board.}
\label{fig:suppl_people}
\end{figure}

\noindent\textbf{Inserted people.}
\cref{fig:suppl_people} relates the camera height implied by photorealistic people painted into set F (classroom and pavilion) to the focal length.
The implied height is correct at 16--24\,mm ($\times$1.0--1.2) and grows to $\times$2.2--3.8 at 85\,mm (log--log slope $+0.55$, $r = 0.88$).
\kontext{} behaves alike ($+0.63$), and so does the neutral floor ($+0.70$).
With the camera read from the painted board instead of the true one, the slope is $+0.42$.
This is the direction of the board's telephoto prior (85\,mm read as about 58\,mm, a factor of 1.46), but the effect is stronger, so generated people are too small for long lenses.

\section{SR-RAW: Optical Zoom and Rotation}
\label{sec:suppl_srraw}

\begin{figure*}[t]
\centering
\includegraphics[width=\textwidth]{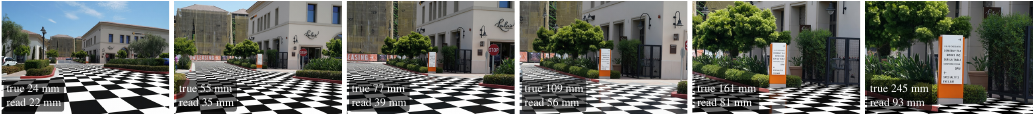}
\caption{\textbf{An SR-RAW sequence painted at six focal lengths} (photographs from SR-RAW~\cite{zhang2019zoom}, \qwen{} with \preplace, variant closest to the median). Titles give the true and the read-out focal length. The painted board widens less than the lens: the 240\,mm frame is painted like a 93\,mm one.}
\label{fig:suppl_srraw}
\end{figure*}

\noindent\textbf{Data.}
The SR-RAW test set~\cite{zhang2019zoom} contains 50 sequences of six or seven camera JPEGs ($4240\times2832$\,px) taken with a Sony FE 24--240\,mm lens on an $\alpha$7S\,II at f/16 to f/36.
Before any editing, we selected on contact sheets the sequences in which a horizontal floor or ground plane (floor, pavement, street, square or flat lawn, but no table or wall) covers about a tenth of the frame or more and is seen obliquely, at the shortest and at least two longer focal lengths.
This leaves 36 sequences, 35 of them outdoors, with 196 frames: 36 each at 24, 35 and 50\,mm, 32 at 70, 24 at 100, 20 at 150 and 12 at 240\,mm (nominal).

\noindent\textbf{Ground truth.}
Registering neighbouring frames with SIFT features and a homography leaves a median residual of 0.63\,px on the frames as delivered and of 1.10\,px after correcting the lens's distortion profile (lensfun), which helps in only 9\% of 162 pairs.
The camera has already corrected the distortion, so we use the frames as they are.
The focal-length ratio of each frame to the widest one is the scale of the chained homographies at the image centre.
It deviates from the EXIF ratio by 2.3\% in the median and 6.0\% at most.
The widest frame is anchored at its EXIF focal length and the sensor width of 35.6\,mm, which gives true focal lengths of 24.2 to 245.3\,mm (36\,mm equivalent).
For every longer frame, a crop of the widest frame with the same field of view serves as its digital twin (160 frames), and all inputs are resized to $1152\times768$\,px.
On synthetic sequences built from the focal sweep of the catalog, with and without artificial lens distortion, this procedure recognises the distortion state and recovers the focal ratios to 0.2\%.
\qwen{} paints two variants per frame with each prompt and \kontext{} one, and GeoCalib and MoGe-2 run on the same inputs.

\begin{table*}[t]
\centering
\caption{\textbf{Read-out focal length per focal length} on the optical SR-RAW frames (median in mm, 35\,mm equivalent). Parentheses give the frames that yield a focal length and all frames. First row: true focal length (median). Painted and exact boards with vanishing points within 10 read-out focal lengths.}
\label{tab:suppl_srraw}
\footnotesize
\setlength{\tabcolsep}{4pt}
\begin{tabular}{@{}lccccccc@{}}
\toprule
Nominal focal length (mm) & 24 & 35 & 50 & 70 & 100 & 150 & 240 \\
\midrule
True focal length & 24 & 36 & 53 & 74 & 106 & 159 & 243 \\
\midrule
Exact synthetic board & 24\,{\scriptsize(36/36)} & 36\,{\scriptsize(35/36)} & 53\,{\scriptsize(36/36)} & 74\,{\scriptsize(31/32)} & 105\,{\scriptsize(22/24)} & 159\,{\scriptsize(19/20)} & 227\,{\scriptsize(8/12)} \\
\qwen, \preplace & 23\,{\scriptsize(23/36)} & 31\,{\scriptsize(25/36)} & 36\,{\scriptsize(21/36)} & 49\,{\scriptsize(21/32)} & 56\,{\scriptsize(12/24)} & 74\,{\scriptsize(12/20)} & 106\,{\scriptsize(6/12)} \\
\qwen, \pcover & 22\,{\scriptsize(22/36)} & 29\,{\scriptsize(21/36)} & 45\,{\scriptsize(24/36)} & 44\,{\scriptsize(17/32)} & 50\,{\scriptsize(11/24)} & 132\,{\scriptsize(8/20)} & 71\,{\scriptsize(6/12)} \\
\kontext, \pcover & 22\,{\scriptsize(15/36)} & 21\,{\scriptsize(17/36)} & 32\,{\scriptsize(23/36)} & 33\,{\scriptsize(9/32)} & 33\,{\scriptsize(13/24)} & 56\,{\scriptsize(9/20)} & 34\,{\scriptsize(7/12)} \\
GeoCalib~\cite{veicht2024geocalib} & 24\,{\scriptsize(36/36)} & 33\,{\scriptsize(36/36)} & 42\,{\scriptsize(36/36)} & 47\,{\scriptsize(32/32)} & 51\,{\scriptsize(24/24)} & 51\,{\scriptsize(20/20)} & 52\,{\scriptsize(12/12)} \\
MoGe-2~\cite{wang2025moge2} & 22\,{\scriptsize(36/36)} & 28\,{\scriptsize(36/36)} & 34\,{\scriptsize(36/36)} & 39\,{\scriptsize(32/32)} & 42\,{\scriptsize(24/24)} & 41\,{\scriptsize(20/20)} & 42\,{\scriptsize(12/12)} \\
 
\bottomrule
\end{tabular}
\end{table*}

\noindent\textbf{Results.}
\cref{tab:suppl_srraw} lists the read-out per focal length (example in \cref{fig:suppl_srraw}).
The painted perspective is slightly too wide already at 24\,mm (22--23\,mm) and falls increasingly short beyond 100\,mm.
GeoCalib is exact at 24\,mm but levels off at about 50\,mm, and MoGe-2 at about 42\,mm.
The probe yields a focal length for 61\% of the frames with \preplace{} and 56\% with \pcover{}, while GeoCalib and MoGe-2 always return one.
Optical and digital zoom to the same field of view are read within $-2$\%\ci{$-6$\%}{$+6$\%} of each other with \preplace, $+1$\%\ci{$-9$\%}{$+9$\%} with \pcover{} and $-2$\%\ci{$-3$\%}{$-1$\%} by GeoCalib, whereas MoGe-2 reads the optical frames 13\% shorter\ci{8\%}{16\%}.
Its slope is 0.65 on the digital twins but 0.36 on the optical frames (paired $-0.37$\ci{$-0.41$}{$-0.34$}).
\kontext{} reads them within $-4$\%\ci{$-20$\%}{$+19$\%} of each other (45 pairs).
The single indoor sequence does not allow an indoor--outdoor comparison.

\begin{table*}[t]
\centering
\caption{\textbf{Bound on the vanishing points}: focal slope (log--log Theil--Sen with 95\% interval, over images with a focal length) when both vanishing points must lie within 10 image heights of the centre (default for catalog and NYUv2), within 10 read-out focal lengths (default for SR-RAW), or anywhere. $^\dagger$Classroom and pavilion only. $^\ast$First variant per image.}
\label{tab:suppl_fp}
\footnotesize
\setlength{\tabcolsep}{4pt}
\begin{tabular}{@{}llccc@{}}
\toprule
Data & Probe & $\le$\,10 image heights & $\le$\,10 read-out focal lengths & no bound \\
\midrule
Catalog renders & \qwen, \pcover & 0.81\,{\scriptsize[0.75,0.87]}\,{\scriptsize(105)} & 0.80\,{\scriptsize[0.74,0.87]}\,{\scriptsize(108)} & 0.79\,{\scriptsize[0.72,0.85]}\,{\scriptsize(119)} \\
 & \kontext, \pcover$^\dagger$ & 0.43\,{\scriptsize[0.27,0.66]}\,{\scriptsize(44)} & 0.38\,{\scriptsize[0.20,0.62]}\,{\scriptsize(45)} & 0.25\,{\scriptsize[0.04,0.52]}\,{\scriptsize(53)} \\
NYUv2 centre crops & \qwen, \pcover & 0.23\,{\scriptsize[0.09,0.34]}\,{\scriptsize(104)} & 0.23\,{\scriptsize[0.12,0.37]}\,{\scriptsize(111)} & 0.30\,{\scriptsize[0.17,0.43]}\,{\scriptsize(151)} \\
 & \qwen, \preplace$^\ast$ & 0.55\,{\scriptsize[0.44,0.66]}\,{\scriptsize(97)} & 0.55\,{\scriptsize[0.44,0.65]}\,{\scriptsize(107)} & 0.54\,{\scriptsize[0.43,0.65]}\,{\scriptsize(132)} \\
SR-RAW optical zoom & \qwen, \preplace & 0.56\,{\scriptsize[0.45,0.65]}\,{\scriptsize(95)} & 0.62\,{\scriptsize[0.52,0.73]}\,{\scriptsize(120)} & 0.69\,{\scriptsize[0.56,0.82]}\,{\scriptsize(167)} \\
 & \qwen, \pcover & 0.41\,{\scriptsize[0.28,0.54]}\,{\scriptsize(92)} & 0.52\,{\scriptsize[0.37,0.73]}\,{\scriptsize(109)} & 0.49\,{\scriptsize[0.32,0.68]}\,{\scriptsize(167)} \\
 & \kontext, \pcover & 0.33\,{\scriptsize[0.17,0.54]}\,{\scriptsize(85)} & 0.36\,{\scriptsize[0.18,0.57]}\,{\scriptsize(93)} & 0.35\,{\scriptsize[0.18,0.53]}\,{\scriptsize(135)} \\
 & exact synthetic board & 1.00\,{\scriptsize[1.00,1.00]}\,{\scriptsize(145)} & 1.00\,{\scriptsize[0.99,1.00]}\,{\scriptsize(187)} & 1.00\,{\scriptsize[1.00,1.00]}\,{\scriptsize(190)} \\
 
\bottomrule
\end{tabular}
\end{table*}

\noindent\textbf{Bound on the vanishing points.}
\cref{tab:suppl_fp} compares three bounds on the vanishing points from which the focal length is computed.
The bound of 10 image heights used for the catalog and NYUv2 also caps readable focal lengths at 10 image heights, 240\,mm for these frames.
To separate what the read-out can do from what the editor paints, we project an exact board of $12\times12$ tiles into every optical SR-RAW frame with its true focal length and read it with the identical pipeline.
The virtual camera is pitched down so that the horizon lies at least a quarter of the image height above the centre, plus 2--15\degr, and the board is rotated by 5--85\degr{} against the view direction.
The exact board passes the image-height bound in only 1 of 12 frames at 240\,mm and 6 of 20 at 150\,mm, but the scale-free bound in all frames in which it is read at all (8 of 12 and 19 of 20), and it is read with slope 1.00 under every bound (240\,mm as 227\,mm).
Because the image-height bound discards correct paintings at long focal lengths, it biases the editor's readable paintings towards too wide perspectives.
For SR-RAW we therefore bound the vanishing points at 10 read-out focal lengths, which requires both line families to be inclined by at least 5.7\degr{} to the image plane.
On the catalog and NYUv2, whose focal lengths stay below 4 image heights, this changes no focal slope by more than 0.05.
Without any bound, estimates from nearly parallel families enter.
The SR-RAW slope with \preplace{} then rises to 0.69 and the NYUv2 slope with \pcover{} to 0.30, the \kontext{} slope on the catalog drops to 0.25, all other slopes change by at most 0.03, and all confidence intervals of the painted boards stay below 1.

\noindent\textbf{Rotation.}
On the rotated frames (\cref{sec:suppl_catalog}, with protocol and expected outcome fixed before the runs), all 720 \qwen{} variants (two per frame and prompt) yield a roll, and GeoCalib runs on the same inputs.
Over all variants, the roll slope is 0.69\ci{0.59}{0.77} with \pcover{} and 0.78\ci{0.69}{0.85} with \preplace{} (intervals over sequences, paired difference $+0.09$\ci{$-0.01$}{$+0.20$}), and 1.00\ci{0.99}{1.01} for GeoCalib.
Roll changes of $-12$\degr{} and $+12$\degr{} are read as $-8.9$\degr{} and $+8.9$\degr{} with \pcover, $-9.5$\degr{} and $+10.6$\degr{} with \preplace{} and $-11.8$\degr{} and $+12.1$\degr{} by GeoCalib.
Both prompts stay below GeoCalib on identical frames (paired $-0.22$\ci{$-0.31$}{$-0.14$} with \preplace{} and $-0.30$\ci{$-0.41$}{$-0.23$} with \pcover).
On the unrotated frames, whose true roll is unknown, GeoCalib measures a median absolute roll of 2.1\degr{} and the probe 1.4\degr{} (\preplace) and 1.5\degr{} (\pcover), as expected from a compression towards level.

\section{Further Editors}
\label{sec:suppl_editors}
We added two open editors, run in ComfyUI with the settings of their model cards and one-megapixel inputs: LongCat-Image-Edit~\cite{longcat2025image} (50 steps, guidance 4.5) and OmniGen2~\cite{wu2025omnigen2} (50 steps, text guidance 5, image guidance 2), both with one variant per image.
In a smoke test on five catalog renders, \longcat{} and OmniGen2 each painted a readable board into four.
For lack of compute time, OmniGen2 was then run only for the default camera (\cref{tab:suppl_default}).
On the catalog (F\,$\cup$\,Z, neutral floor), \longcat{} paints a readable board into 89 of the 120 renders, 58 of them registrable (\cref{sec:suppl_readout}).
These do not differ from the others in true roll (median 5.6\degr{} \vs 5.7\degr, $p = 0.66$, Mann--Whitney) or focal length (31\,mm both, $p = 0.58$).
On them, \longcat{} shows both priors about as strongly as \kontext{} (\cref{tab:priors}), with a roll slope of 0.25\ci{0.13}{0.36}, a focal slope of 0.34\ci{0.22}{0.51} and a fixed point of 28\,mm.
Unlike \qwen, it also compresses pitch slightly (0.84\ci{0.66}{0.94}).
It is the least accurate of the three editors in horizon and pitch (\cref{tab:accuracy}).

\section{Writing versus Reading: Details}
\begin{figure}[t]
\centering
\includegraphics[width=0.8\spalte]{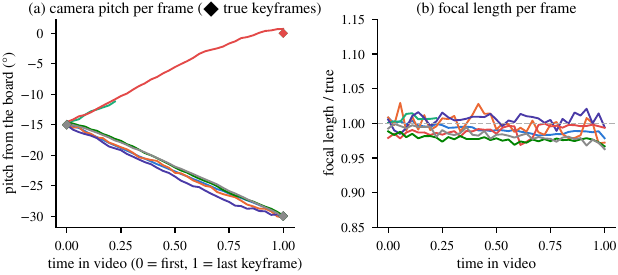}
\caption{\textbf{Camera trajectories of generated videos}, read frame by frame from the board (LTX-2.5 between two renders). Diamonds mark the true keyframe pitch.}
\label{fig:suppl_video}
\end{figure}
\label{sec:suppl_writeread}

\begin{table}[t]
\centering
\caption{\textbf{Multi-angle LoRA}: median output pitch (\degr) produced (GeoCalib) and believed (painted board), their median absolute difference $\Delta$, and focal length output/input by GeoCalib (GC) and by the board. Inputs have a median pitch of $-15$\degr{} ($-11$\degr{} to $-22$\degr). For eye-level, elevated, high-angle and front-right, 12 inputs at 0\degr{} and $-30$\degr{} are included. Variants with a failed registration are discarded. Reference on the original renders: $\Delta = 2.0$\degr.}
\label{tab:suppl_lora}
\small
\setlength{\tabcolsep}{2.8pt}
\begin{tabular}{@{}lrrrrrr@{}}
\toprule
Command (card) & $n$ & produced & believed & $\Delta$ & $f$ GC & $f$ board \\
\midrule
eye-level (0\degr) & 25 & $-9.9$ & $-12.0$ & 1.4 & 0.99 & 1.02 \\
elevated (30\degr) & 26 & $-26.6$ & $-26.7$ & 2.4 & 0.96 & 0.97 \\
high-angle (60\degr) & 25 & $-36.9$ & $-37.1$ & 3.4 & 0.89 & 0.98 \\
low-angle ($-30$\degr) & 15 & $+2.7$ & $+3.5$ & 2.0 & 0.98 & 0.99 \\
close-up ($\times$0.6) & 14 & $-12.7$ & $-14.7$ & 1.5 & 1.01 & 1.02 \\
wide ($\times$1.8) & 16 & $-1.7$ & $-0.2$ & 1.5 & 0.86 & 0.82 \\
front-right (45\degr) & 27 & $-10.3$ & $-10.6$ & 1.4 & 0.90 & 0.98 \\
\bottomrule
\end{tabular}
\end{table}

\cref{tab:suppl_lora} lists all commands.
Theil--Sen fits of output on input pitch (inputs at 0\degr, $-15$\degr{} and $-30$\degr) give $0.72\times\text{input} - 0.8$\degr{} (eye-level), $0.44\times\text{input} - 20.1$\degr{} (elevated) and $0.59\times\text{input} - 28.4$\degr{} (high-angle), between an absolute target pitch (slope 0) and a change relative to the input (slope 1).
``Wide shot'' opens the field of view and levels the camera, while ``close-up'' hardly changes a room.
The yaw change of ``front-right'' is the board's yaw in the output minus the true yaw of the input (accuracy of the board's yaw on the neutral-floor renders: 1.1\degr{} median error and 83\% within 5\degr, against 0.1\degr{} and 98\% for the rendered board).
On renders, 22 of 24 front-right views turn in the same direction, by 30\degr{} in the median (interquartile range 22--36\degr), while the eye-level control stays at $-0.1$\degr{} ($-2.1$ to $+0.7$\degr).
On 20--40 NYUv2 photographs ``elevated'' lowers the pitch by 17\degr{} relative to eye-level (renders: 15\degr{} at an input pitch of $-15$\degr), and of 36 measurable front-right views (input yaw also read from a board) 21 turn by about $+33$\degr, 10 by about $-41$\degr{} (modulo 90\degr{} indistinguishable from $+49$\degr, but a mirrored turn is more plausible), 4 by $-5$\degr{} to $-25$\degr{} and one not at all.

\noindent\textbf{Video.}
\cref{fig:suppl_video} shows the seven LTX-2.5 videos between two renders with a physical board (73 frames, every second one measured).
All trajectories are monotonic up to single backward steps of at most 0.10\degr{} (two videos) and start within 0.4\degr{} and end within 0.7\degr{} of the true keyframes.
One video ending at a nearly level camera is measurable in only 22\% of the frames (board too flat, last reading $-11.2$\degr).
The focal length stays at 0.98--1.01 of the truth (spread 0.4--1.4\%), roll at most 1.0\degr.
Towards an ``elevated'' LoRA view (four videos) the pitch moves monotonically in 77--100\% of the steps, with more focal spread (1--6\%) because the keyframes carry differently painted boards.
Towards a front-right view, the pitch moves monotonically in 82--100\% of the steps, and the focal spread is larger (2--32\%).

\section{Reproducibility}
\label{sec:suppl_repro}
All numbers are computed by scripts from per-image measurements on the lossless editor outputs.
Both main tables and all figures are generated automatically, and measurements, ground truth, renders and JPEG copies of all outputs are archived.
Editing ran on one consumer GPU (median per variant including read-out: 8\,s for \qwen{} with the Lightning distillation, 51\,s with the full 40-step model, 73\,s for \longcat).
Lightning paints the same perspective as the full model (classroom sweep: focal slopes 0.67 and 0.66, paired focal error not significantly different).
Confidence intervals use 1000 bootstrap resamples over images (2000--4000 for some paired tests), drawn jointly for paired comparisons.

\end{document}